\documentclass[journal]{IEEEtran}

\ifCLASSINFOpdf

\else

\fi

\usepackage[utf8]{inputenc} 
\usepackage[T1]{fontenc}    
\usepackage{url}            
\usepackage{booktabs}       
\usepackage{amsfonts}       
\usepackage{nicefrac}       
\usepackage{microtype}      
\usepackage{lipsum}		
\usepackage{graphicx}
\usepackage{doi}
\usepackage{indentfirst} 
\usepackage{smartdiagram}
\usepackage{subfigure}
\usepackage{multirow}
\usepackage{longtable}
\usepackage{pdflscape}
\usepackage{bbding}
\usepackage{makecell}
\usepackage[figuresright]{rotating}
\usepackage{color,xcolor}
\usepackage{algpseudocode}  
\usepackage{amsmath}  
\usepackage{cite}

\usepackage{amsthm}

\usepackage{hyperref} 

\usepackage{array}
\usepackage{booktabs,makecell, multirow, tabularx}
\usepackage{threeparttable}
\usepackage{longtable}
\usepackage{afterpage}
\usepackage{xcolor}

\usepackage{changes} 
\definechangesauthor[name={Jianqi Gao}, color=red]{1} 
\definechangesauthor[name={Jianqi Gao}, color=blue]{2} %

\makeatletter  
\newif\if@restonecol  
\makeatother

\usepackage[linesnumbered,ruled,vlined]{algorithm2e}
\usepackage{algpseudocode}  

\usepackage{tikz,xcolor,hyperref}
\definecolor{lime}{HTML}{A6CE39}
\DeclareRobustCommand{\orcidicon}{%
	\begin{tikzpicture}
	\draw[lime, fill=lime] (0,0) 
	circle [radius=0.16] 
	node[white] {{\fontfamily{qag}\selectfont \tiny ID}};    \draw[white, fill=white] (-0.0625,0.095) 
	circle [radius=0.007];    \end{tikzpicture}
	\hspace{-2mm}}
\foreach \x in {A, ..., Z}{%
	\expandafter\xdef\csname orcid\x\endcsname{\noexpand\href{https://orcid.org/\csname orcidauthor\x\endcsname}{\noexpand\orcidicon}}
}

\begin{document}
	
\IEEEpubid{\begin{minipage}{\textwidth}\centering
				Copyright \copyright 2025 IEEE. Personal use of this material is permitted. \\
				However, permission to use this material for any other purposes must be obtained from the IEEE by sending an email to pubs-permissions@ieee.org.
			\end{minipage}}

\bibliographystyle{ieeetr}

\title{Hierarchical Reinforcement Learning for Safe Mapless Navigation with Congestion Estimation}

\newcommand{\orcidauthorA}{0000-0003-4486-3740} 
\newcommand{\orcidauthorB}{0000-0003-0322-5590}
\newcommand{\orcidauthorC}{0000-0002-7767-8157}
\newcommand{\orcidauthorD}{0000-0001-7860-9666}

\author{Jianqi Gao, Xizheng Pang, Qi Liu, and Yanjie~Li, ~\IEEEmembership{Member,~IEEE}

\IEEEcompsocitemizethanks
	{
	\IEEEcompsocthanksitem 
	This work was supported
	in part 
	by the Shenzhen Fundamental Research Program under Grant JCYJ20220818102415033, JSGG20201103093802006 and  KJZD20230923114222045 and 
	in part 
	by the National Natural Science Foundation of China under Grant 61977019 and  U1813206. (\textit{Jianqi Gao and Xizheng Pang contribute equally. Corresponding author: Yanjie Li and Qi Liu}.)
	\IEEEcompsocthanksitem
	Jianqi Gao, Xizheng Pang, and Yanjie Li
	are with the Department of Guangdong Key Laboratory of Intelligent Morphing Mechanisms and Adaptive Robotics and 
	School of Inteligence Science and Engineering,
	Harbin Institute of Technology, Shenzhen, 518055, China.
	{\tt\small gaojianqi205a@stu.hit.edu.cn, xzpang@stu.hit.edu.cn, autolyj@hit.edu.cn}
	\IEEEcompsocthanksitem
	Qi Liu is with the Faculty of Robot Science and Engineering, Northeastern University, 110819, Shenyang, China.
	{\tt\small liuqi@mail.neu.edu.cn}
	\IEEEcompsocthanksitem
	\textbf{Copyright \copyright 2025 IEEE. This paper has been accepted by ICRA2025. Personal use of this material is permitted. 
	However, permission to use this material for any other purposes must be obtained from the IEEE by sending an email to pubs-permissions@ieee.org.}
	\IEEEcompsocthanksitem
	Citation format: J. Gao, X. Pang, Q. Liu, Y. Li, "Hierarchical Reinforcement Learning for Safe Mapless Navigation with Congestion Estimation", 2025 IEEE International Conference on Robotics and Automation (ICRA), 2025.
	}
}


\maketitle

\pagestyle{empty}
\thispagestyle{empty}


\begin{abstract}
Reinforcement learning-based mapless navigation holds significant potential. However, it faces challenges in indoor environments with local minima area. This paper introduces a safe mapless navigation framework utilizing hierarchical reinforcement learning (HRL) to enhance navigation through such areas. The high-level policy creates a sub-goal to direct the navigation process. Notably, we have developed a sub-goal update mechanism that considers environment congestion, efficiently avoiding the entrapment of the robot in local minimum areas.
The low-level motion planning policy, trained through safe reinforcement learning, outputs real-time control instructions based on acquired sub-goal. Specifically, to enhance the robot's environmental perception, we introduce a new obstacle encoding method that evaluates the impact of obstacles on the robot's motion planning.
To validate the performance of our HRL-based navigation framework, we conduct simulations in office, home, and restaurant environments. The findings demonstrate that our HRL-based navigation framework excels in both static and dynamic scenarios. Finally, we implement the HRL-based navigation framework on a TurtleBot3 robot for physical validation experiments, which exhibits its strong generalization capabilities.
\end{abstract}



\IEEEpeerreviewmaketitle

\section{Introduction}

\IEEEPARstart{N}{avigation} is among the most fundamental and critical functionalities of robots \cite{gul2019comprehensive,el2021indoor,loganathan2023systematic}. While traditional navigation algorithms are adept at ensuring that robots safely and efficiently accomplish their designated tasks, they often encounter difficulties when adapting to environments that are mapless, dynamic and unstructured \cite{dong2023review}. Although mapless navigation algorithms, such as the Bug algorithm \cite{mcguire2019comparative,sanchez2021path,patil2024comprehensive}, mitigate some of these challenges, they tend to produce excessively long paths

In recent years, researchers have acknowledged that reinforcement learning provides robots with the ability to autonomously learn and adapt in unfamiliar environments, offering significant advantages in generalization and real-time decision-making \cite{Lei2017IROS,Marchesini2020ICRA,zhu2021deep,singh2023review}. However, reinforcement learning-based mapless navigation faces challenges such as robots frequently encountering local minimum areas \cite{xu1999sensor,velagic2006efficient,melchiorre2023experiments} that hinder task completion, a tendency to take detours \cite{liu2023graph} due to the low likelihood of selecting direct actions from the neural network's output, and the lack of effective safety guarantees, potentially leading to collisions \cite{brunke2022safe}.

\begin{figure}[t]
	\centering
	\subfigure[Environment.]
	{
		\label{fig:path-1}
		\includegraphics[width=0.465\linewidth]{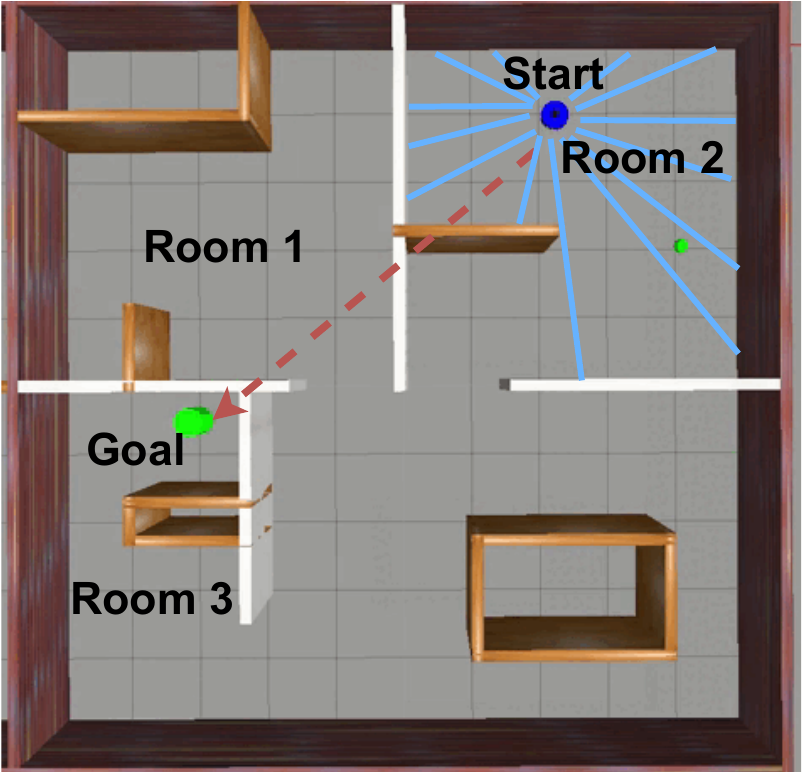}
	}
	\hspace{-2.5mm}
	\subfigure[Navigation path with sub-goal.]
	{
		\label{fig:path-2}
		\includegraphics[width=0.465\linewidth]{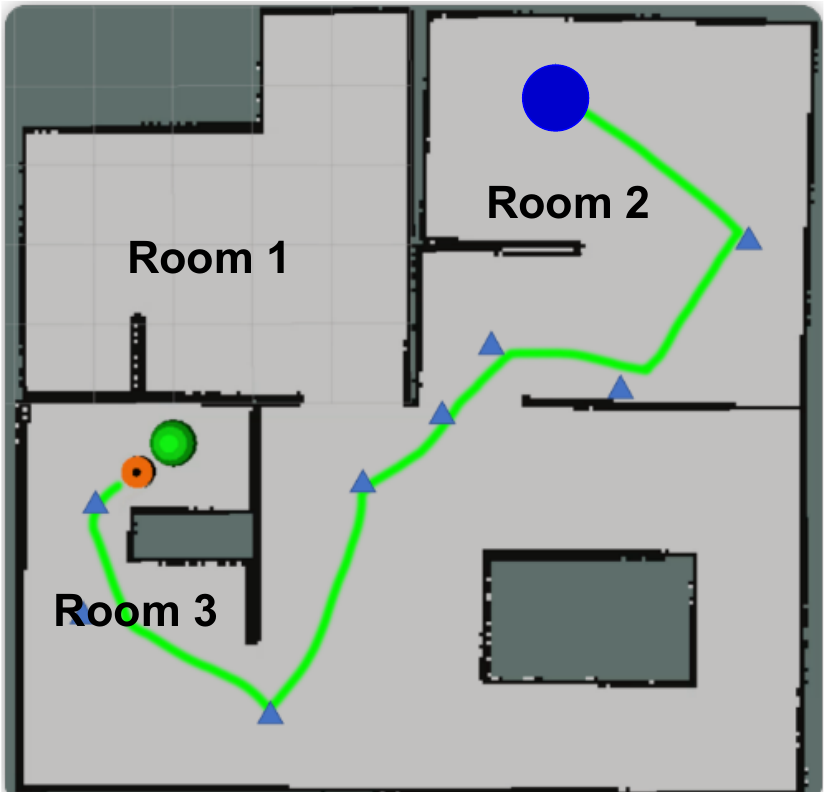}
	}
	\caption{An illustration of HRL-based safe mapless navigation. 
	In (a), the blue dot in room 2 marks the robot's start, while the green dot in room 3 indicates the goal. The red dotted arrow shows the direction from start to goal, and the blue line is the laser scan from the lidar. Lacking a global map, the robot relies on relative goal information and may follow the red arrow, potentially trapping it in room 2. In (b), the green curve is the navigation path with the blue triangle as a sub-goal, and the red dot marks the robot's current position.
	}
	\label{fig:Mapless_Navigation}
	\vspace{-1em}
\end{figure}

\textit{Contribution}:
As depicted in Fig. \ref{fig:Mapless_Navigation}, this paper introduces a mapless navigation framework leveraging hierarchical reinforcement learning (HRL) to address the aforementioned challenges. The high-level policy formulates sub-goals that direct the robot's navigation.
Specifically, we have developed a sub-goal update mechanism that accounts for environment congestion, 
effectively preventing the robot from falling into local minimum areas. Concurrently, the low-level motion planning policy, honed through safe reinforcement learning (SRL), produces real-time control directives utilizing the achieved sub-goal, lidar-derived obstacle data, and current robot status. To enhance the robot's environmental awareness, we've implemented an innovative obstacle encoding technique that quantifies obstacles' effects on motion planning.
To evaluate the navigation capabilities of the proposed HRL-based framework, we simulate indoor environments, including office, home, and restaurant, within Gazebo and performe physical tests using the TurtleBot3-Burger robot. The results demonstrate that the HRL-based framework excels across various settings, showcasing its ability to navigate through local minimum areas during indoor navigation.

The remainder of this paper is structured as follows.
Section \ref{sec:Problem_Formulation} presents the problem formulation. Section \ref{sec:HRL-Based_Navigation} introduces the HRL-based navigation framework.
Section \ref{sec:Simulation_Experiments} presents the simulation experiment results. Section \ref{sec:Validation_Experiments} shows the physical robots validation. 
Section \ref{sec:conclusions} provides the conclusion.

\section{Problem Formulation}
\label{sec:Problem_Formulation}


To simplify and formalize the description, the robot's navigation can be framed as a finite-horizon Partially Observable Markov Decision Process (POMDP), denoted by the tuple \( \langle h, S, A, \Omega, T, O, R, \gamma, b_{0} \rangle \). Here, \( h \) represents the planning horizon, while \( S \), \( A \), and \( \Omega \) denote the sets of states, actions, and observations, respectively. The transition model \( T \) is stochastic, with \( T(s^t, a^t, s^{t+1}) = P(s^{t+1} \mid s^t, a^t) \) indicating the probability of transitioning to state \( s^{t+1} \) from state \( s^t \) after taking action \( a^t \). The observation model \( O \) is given by \( O(o^{t+1}, s^{t+1}, a^t) = P(o^{t+1} \mid s^{t+1}, a^t) \), representing the likelihood of observing \( o^{t+1} \) when reaching state \( s^{t+1} \) following action \( a^{t} \). The reward function \( R \) is bounded, where \( R(s^t, a^t) \) is the reward received for performing action \( a^t \) in state \( s^t \). The discount factor \( \gamma \), ranging from 0 to 1, balances the importance of immediate versus future rewards. Lastly, \( b_{0} \) is the initial state distribution, with \( b_{0}(s^t) \) equal to the probability of the system's initial state being \( s^t \).

\begin{figure}[ht]
	\centering
	\includegraphics[width=0.9\linewidth]{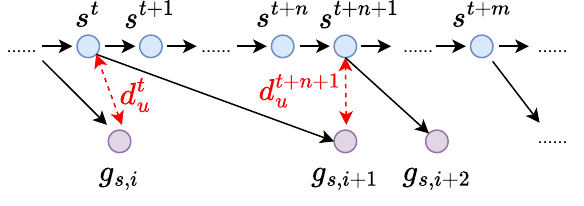}
	\caption{The workflow of the proposed HRL-based navigation framework.}
	\label{fig:image_HER}
\end{figure}

As depicted in Fig. \ref{fig:image_HER}, we address the POMDP using a hierarchical framework. The high-level policy generates sub-goals, while the low-level policy, conditioned on these sub-goals, produces executable actions. The high-level policy activates when the robot is within a distance \( d^t_u \) (defined in (\ref{e:update_distance})) from the current sub-goal. With a given sub-goal, the low-level policy determines actions at each time step to guide the robot towards the new sub-goal while evading both static and dynamic obstacles. Formally, the low-level sub-goal-conditioned policy is denoted by $ \pi\left(a^{t} \mid {s}^{t}, {g}_{s,i}\right) $, where ${g}_{s,i}$ is the new $i$th sub-goal.
The low-level policy essentially solves a short-horizon goal-reaching POMDP defined between two consecutive high-level sub-goals. 

\section{HRL-Based Navigation Framework}
\label{sec:HRL-Based_Navigation}

\subsection{Overview}

\begin{figure}[t]
	\centering
	\includegraphics[width=\linewidth]{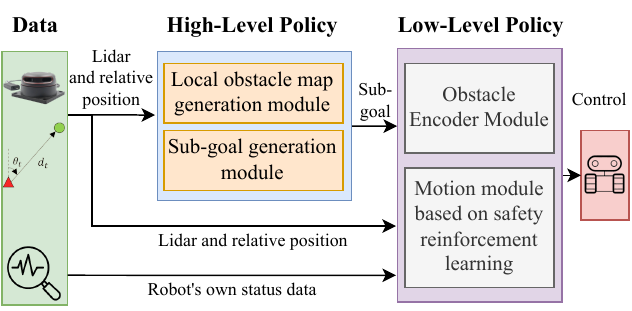}
	\caption{Overall of the HRL-based safe mapless navigation policy.}
	\label{fig:framework_HRL-based_navigation}
\end{figure}

The proposed HRL-based navigation framework is depicted in Fig. \ref{fig:framework_HRL-based_navigation}. The high-level policy uses environment data to generate sub-goals within the robot's lidar range, guiding motion planning and avoiding local minima areas. The low-level policy issues real-time control commands based on environment, robot status, and sub-goal data, allowing for agile obstacle and pedestrian avoidance. Due to their independence, both policies can be trained and deployed separately.



\subsection{High-Level Sub-Goal Policy}





\subsubsection{Observation Space}

%
%


The high-level policy network's observation space integrates lidar point cloud data \( \mathbb{L}^t = [l^t_1, l^t_2, l^t_3, \ldots, l^t_n] \) ($n$ is 1080, and the detection distance is 0.3 to 6 meters) and the goal's relative polar coordinates $\left [ d^t,\theta^t \right ] $. We preprocess the raw lidar data into a local obstacle map (LOMap), which we then input into the network alongside maps from the previous three timesteps. This method provides a compact scene representation, enhancing the robot's ability to differentiate between similar scenes and improving its historical memory for escaping local minima aera. (LOMap details in supplementary material.)

\begin{figure}[ht]
	\centering
	\subfigure[Environment.]
	{
		\label{fig:LOM-1}
		\includegraphics[width=0.325\linewidth]{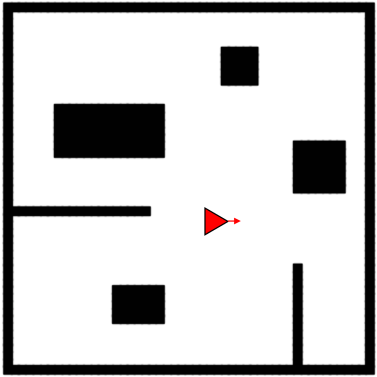}
	}
	\hspace{-2.5mm}
	\subfigure[Local obstacle map.]
	{
		\label{fig:LOM-2}
		\includegraphics[width=0.35\linewidth]{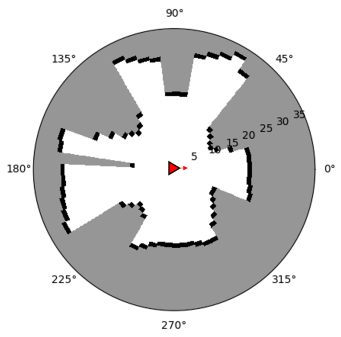}
	}
	\caption{An illustration of local obstacle map. The red triangle denotes the robot, with its arrow indicating the robot's direction. The unoccupied areas are depicted in white, the occupied areas in black, and the unexplored areas in gray.}
	\label{fig:image_LOM}
	\vspace{-1em}
\end{figure}

\subsubsection{Action Space}

\begin{figure}[ht]
	\centering
	\subfigure[]
	{
		\label{fig:image_high-level_action_space-1}
		\includegraphics[width=0.35\linewidth]{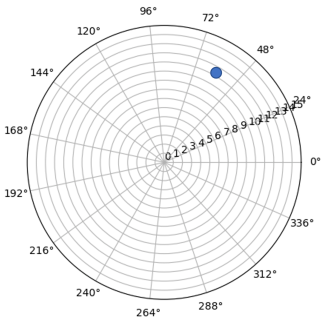}
	}
	\hspace{-2.5mm}
	\subfigure[]
	{
		\label{fig:image_high-level_action_space-2}
		\includegraphics[width=0.35\linewidth]{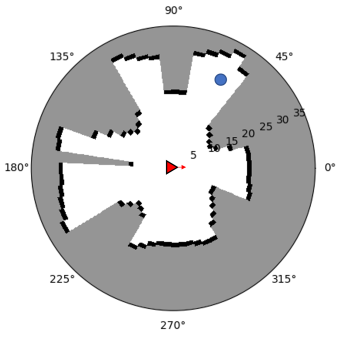}
	}
	\caption{The polar coordinate of the sub-goal. We discretize the detectable range of the robot's lidar into 15 equal distance intervals and 360° into 15 equal angles, thereby discretizing the local obstacle map into 225 sector-shaped areas. The blue dot represents a sub-goal.}
	\label{fig:image_high-level_action_space}
	\vspace{-1em}
\end{figure}

As shown in Fig. \ref{fig:image_high-level_action_space}, the action space of the high-level policy consists of the polar coordinates of the sub-goal relative to the current position of the robot, denoted as \({g}_{s,i}= [d_{{sg},i}, \theta_{{sg},i}] \), where \(d_{{sg},i}\) is in meters and \(\theta_{{sg},i}\) is in radians. The sub-goal coordinates are constrained to the unoccupied areas within the generated local obstacle map.
For areas that are inaccessible to the robot, we apply an action mask operation to the output of the policy network.


\subsubsection{Reward Function}
To encourage the robot to reach its goal promptly, the reward function is designed as follows:
\begin{equation}
	r=r_{\text {arrival }}+r_{\text {step }}+r_{\text {dist }}+r_{\text {out }},
	\label{e:high-level_totoal_reward}
\end{equation}
where $r_{\text {arrival }}$ represents the reward for reaching the goal, whichc can be defined as:
\begin{equation}
	r_{\text {arrival }}=\left\{\begin{array}{l}
		100, d_{t}<d_{\text {limit }} \\
		0, \quad \text { others }
	\end{array}\right.,
	\label{e:high-level_reward_arrival}
\end{equation}
when the robot's distance to the goal \( d_t \) is less than the threshold \( d_{\text{limit}} \) (0.5 meters in this paper), the robot is considered to have reached the goal, and the reward \( r_{\text {arrival }} \) is 100; otherwise, \( r_{\text {arrival }} \) is 0.
The value \( r_{\text{step}}  \)= -1 indicates a penalty of -1 for each sub-goal selection, encouraging the robot to minimize detours and reach the goal promptly, aiming for an optimal path.
The reward \( r_{\text{dist}} \) encourages the robot to approach the goal, calculated as the difference in distances to the goal between consecutive timesteps using the A\(^*\) algorithm on a \(0.5 \times 0.5\) grid map:\footnote{During training, the high-level policy network uses A$^*$-based distances, eliminating the need for A$^*$ distance calculations during actual policy deployment, thus satisfying mapless navigation requirements.}
\begin{equation}
	r_{d i s t}=\mu\left(d^{t-1}_{A *}-d^{t}_{A *}\right),
	\label{e:high-level_reward_dist}
\end{equation}
where the coefficient \( \mu = 1 \) in this paper.
The penalty reward \( r_{\text{out}} \) = -20 applies when the robot selects an undesirable sub-goal, like one near or blocked by an obstacle.

\subsubsection{Policy Network and Training}

\begin{figure}[ht]
	\centering
	\includegraphics[width=\linewidth]{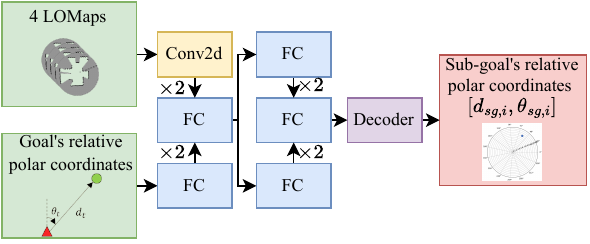}
	\caption{An illustration of high-level policy network. 'Cond2v' for 2D convolutional layers and 'FC' for fully connected layers.}
	\label{fig:image_policy_network}
\end{figure}

The architecture
of the high-level policy network are shown in Fig. \ref{fig:image_policy_network}.
The reinforcement learning algorithm known as dueling double deep Q-networks algorithm \cite{wang2016dueling,van2016deep} is utilized to train the high-level policy network.
Given the sparse reward nature of robot navigation, the weight ratio of rewards can impact both navigation performance and training efficiency, despite reward reshaping. Additionally, the complexity and variability of indoor environments pose training challenges. Thus, we also employ the hindsight experience replay \cite{andrychowicz2017hindsight} method to refine the training process.

\subsubsection{Environment Congestion-Based Sub-Goal Update}

Since path planning requires a certain amount of calculation time, the robot significantly pauses when it reaches a sub-goal and then calculates the next one, resulting in low navigation efficiency. As the robot approaches the current sub-goal, we can generate the subsequent sub-goal, thereby alleviating the aforementioned phenomenon to a certain extent. However, in crowded scenes, with rapid changes in environment observations, setting the sub-goal update distance threshold too large can cause the robot to freeze or wander aimlessly. 


To address these issues, this paper proposes a dynamic update of sub-goals based on environment congestion.
First, we define environment congestion at $t$ as follows:
\begin{equation}
C^t=\frac{1}{n} \sum_{i=1}^{n} \frac{1}{\log _{d_{s}}\left(l^t_{i}+1\right)},
\label{e:env_congest}
\end{equation}
where \( n \) is the dimension of the lidar point cloud data \( \mathbb{L}^t = [l^t_1, l^t_2, l^t_3, \ldots, l^t_n] \), which is 1080. \( d_s \) is the safe distance threshold, set to 3 meters. The larger the value of the environment congestion coefficient \( C^t \), the more crowded the environment is deemed. If the environment congestion degree \( C^t < 1 \), it indicates that the environment is relatively unoccupied.
Next, we define the sub-goal update distance threshold \( d^t_u \). The sub-goal update is triggered when the distance between the robot and the current sub-goal is less than \( d^t_u \):
\begin{equation}
d^t_{u}=\operatorname{clip}(\alpha C^t+\beta, 0.5,2),
\label{e:update_distance}
\end{equation}
where \( \alpha \) is a coefficient related to the robot's speed and maneuverability, taken as 0.25 in this paper, and \( \beta \) is also taken as 0.25. Furthermore, to prevent the robot from lingering in one area, a time threshold for sub-goal update is established at 30 seconds. That is, if the robot does not move within the sub-goal update distance threshold \( d^t_u \) within 30 seconds, a new sub-goal is forcibly regenerated.
The sub-goal update mechanism, based on environment congestion, enables the robot to dynamically update sub-goals in response to surrounding obstacles and pedestrians, effectively preventing the robot from falling into local minimum areas.

\subsection{Low-Level Sub-Goal-Conditioned Motion Planning Policy}

The low-level motion policy is primarily tasked with enabling the robot to avoid both static and dynamic obstacles and must generate control instructions in real time. As depicted in Fig. \ref{fig:framework_HRL-based_navigation}, this policy primarily comprises an obstacle encoder module for preprocessing state information and a motion planning module that utilizes secure reinforcement learning.


	
\subsubsection{Observation Space}




For obstacle avoidance during the robot's movement to the sub-goal, the observation space of the low-level motion planning policy encompasses the lidar point cloud data \( \mathbb{L}^t \), the robot's movement speed \( [v^t, \omega^t] \), and the polar coordinates of the sub-goal \({g}_{s,i}= [d_{{sg},i}, \theta_{{sg},i}] \)) provided by the high-level policy. Considering the robot's inertia, obstacle avoidance policies vary at high and low speeds; hence, the current wheel odometer information is translated into the linear velocity \( v_t \) and angular velocity \( \omega_t \) of robot as part of the observation space.

\subsubsection{Action Space}

The action space consists of the speed control commands for the robot, denoted as \([v^{t+1}, \omega^{t+1}]\), where the linear velocity \(v^t\) is measured in meters per second and the angular velocity \(\omega^t\) is measured in radians per second.

\subsubsection{Reward Function}

We define the reward function as:
\begin{equation}
	r=r_{\text {arrival }}+r_{\text {step }}+r_{\text {dist }},
	\label{e:low-level_reward_total}
\end{equation}
where each part is consistent with (\ref{e:high-level_totoal_reward}).

\subsubsection{Policy Network and Training}
\label{sec:{Policy_Network_Training}}

\begin{figure}[ht]
	\centering
	\includegraphics[width=0.9\linewidth]{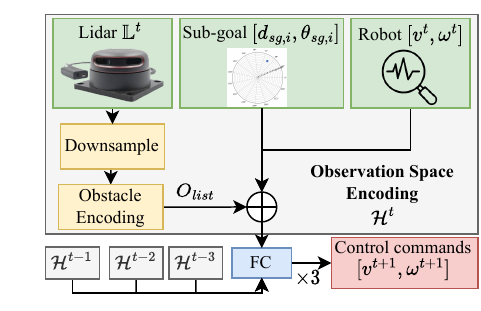}
	\caption{Low-level motion planning Policy network.}
	\label{fig:image_low_level_policy_network_actor}
\end{figure}

The policy actor network architecture is depicted in Fig. \ref{fig:image_low_level_policy_network_actor}. For the lidar point cloud data \( \mathbb{L}^t \), we initially perform minimum pooling downsampling, followed by encoding obstacles within the safety distance threshold \( d_s \) to extract the adjacent obstacle set \( O^t_{list} \) (further details are provided in section \ref{sec:Obstacle_Encoder}).\footnote{Unlike the high-level sub-goal generation policy, the low-level motion planning policy demands faster real-time adjustments for robot safety in dynamic environments, necessitating simple lidar point cloud data preprocessing.}  
The processed lidar data, combined with the sub-goal and the robot's current velocity data, forms the observation space code \( \mathcal{H}_t \). 
Since the robot cannot obtain global information in a mapless navigation, it may end up in a wrong room. In order to solve this problem, \( \mathcal{H}_t \) is concatenated with the \( \mathcal{H}_{t-1},\mathcal{H}_{t-2},\mathcal{H}_{t-3} \) from the previous three time steps, enabling the robot to have historical memory functions. Ultimately, all observation space encoder information are fed into a three-layer fully connected network to determine \( [v^t, \omega^t] \).
Considering the safety of robot navigation, we use the SRL algorithm, constrained policy optimization (CPO) \cite{achiam2017constrained}, to train the policy network.
\footnote{SRL streamlines reward function design and automatically balances the agent's behavior between conservative and aggressive approaches, enhancing the mobile robot's navigation success rate and minimizing collision frequency.}
Unlike typical reward functions, (\ref{e:low-level_reward_total}) omits a collision penalty term, as CPO incorporates collision avoidance within its safety constraints. 

\subsubsection{Obstacle Encoding}
\label{sec:Obstacle_Encoder}


We assume that the closer the obstacle is to the robot, the greater the threat to its navigation, and the closer the obstacle is to the robot's direction of movement, the more dangerous it is. In order to help the robot better perceive obstacles, this paper encodes them according to the threat level of the obstacles.
For the 1080-dimensional single-line lidar point cloud data \( \mathbb{L}^t \), we first remove the noise, and then take the frame with the smallest sampling distance value in every 30 frames of radar data points as the obstacle point cloud data at that angle, and then we get a 36-dimensional sparse radar data. Next, we encode the obstacle point cloud within the safety distance threshold $d_s$ in the format of $ \left[d_{i}^{o,t}, \theta_{i}^{o,t}\right] $ to obtain the adjacent obstacle set $ O^t_{\text {list }} $, where $ d_{i}^{o,t} $, $\theta_{i}^{o,t} $ represent the distance value of the $ i $th obstacle point cloud and the angle of deviation from the positive direction of the robot, respectively. Finally, we sort the elements in $O^t_{\text {list }}$ according to the threat level of the obstacles from small to large, and obtain $ O^t_{\text {list }}=\left(\left[d_{0}^{o,t}, \theta_{0}^{o,t}\right], \cdots,\left[d_{i}^{o,t}, \theta_{i}^{o,t}\right], \cdots,\left[d_{n}^{o,t}, \theta_{n}^{o,t}\right]\right) $, where $ n = 36 $, and for $ i<j $, $ d_{i}^{o,t}>d_{j}^{o,t} $ or $ d_{i}^{o,t}=d_{j}^{o,t}, \theta_{i}^{o,t} \geq \theta_{j}^{o,t} $.

\section{Simulation Experiments}
\label{sec:Simulation_Experiments}

In this section, we present the results of our simulation experiments with the HRL-based navigation framework. 

%
%
%

\subsection{Experiment Settings}

Due to the independence of the high-level and low-level policies, we train the two policy networks separately before deploying them in conjunction. The high-level sub-goal generation policy is initially trained and tested within a 2D Python simulator. Subsequently, for the low-level motion planning policy, we construct a simulation environment utilizing ROS-Kinetic Gazebo 9, wherein the robot model employed is the Turtlebot. To introduce dynamic obstacles into the simulation, we incorporate the social force model-based pedestrians \cite{helbing1995social}.

\subsection{Metrics}

We designed metrics to evaluate navigation performance:

1) Success rate (SR): The ratio of successful goal reaches to total tests. We conduct 100 tests in each scenario.

2) Success Rate of Non-collision (SRN): The ratio of successful, collision-free goal reaches to total tests.

3) Average Collision Times (CT): The average number of collisions per test. Rounds aren't terminated by collisions, so the CT can exceed 1.

4) Success weighted by Path Length (SPL) \cite{anderson2018evaluation}: A measure of navigation path quality, calculated as:
\begin{equation}
S P L=\frac{1}{N} \sum_{i=1}^{N} S_{i} \frac{\tilde{p}_i }{\max \left(p_{i}, \tilde{p}_i \right)},
\label{e:SPL}
\end{equation}
where \( S_i \) is 1 for a successful and 0 for an unsuccessful navigation, \( \tilde{p}_i  \) is the theoretical shortest path, \( p_i \) is the actual path length, and \( N \) is the number of trials. Higher SPL values indicate better navigation.

5) Success weighted by Navigation Time (SNT): A new metric assessing time efficiency, defined as:
\begin{equation}
S N T=\frac{1}{N} \sum_{i=1}^{N} S_{i} \frac{t_{i}}{\max \left(q_{i}, t_{i}\right)},
\label{e:SNA}
\end{equation}
where \( q_i \) is the actual navigation time, $t_i$ is the shortest time for the $i$th navigation, obtained by dividing the optimal distance based on A* by the robot’s maximum speed

%
%

\subsection{Ablation Experiments}

\subsubsection{LOMap for Sub-Goal Generation}

\begin{figure}[ht]
	\centering
	\subfigure[Cumulative reward]
	{
		\label{fig:image_LOMap_training_input_return}
		\includegraphics[width=0.47\linewidth]{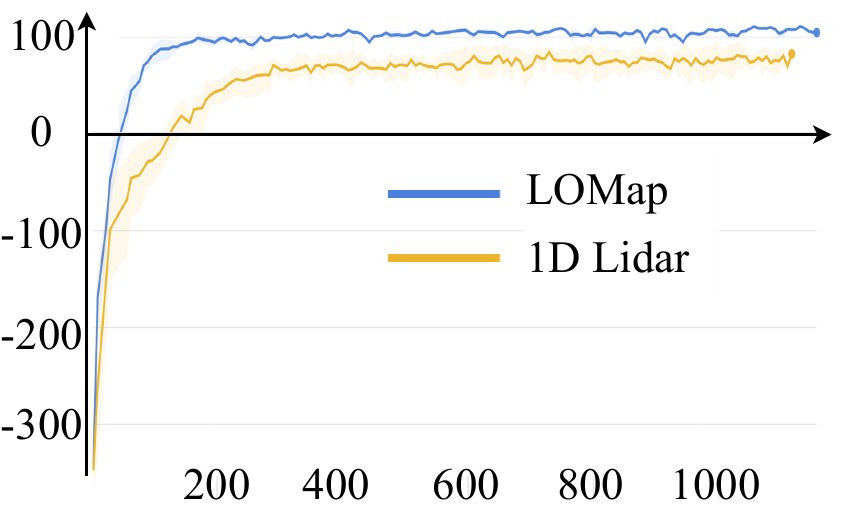}
	}
	\hspace{-2.5mm}
	\subfigure[Average number of steps per round]
	{
		\label{fig:image_LOMap_training_input_loss}
		\includegraphics[width=0.47\linewidth]{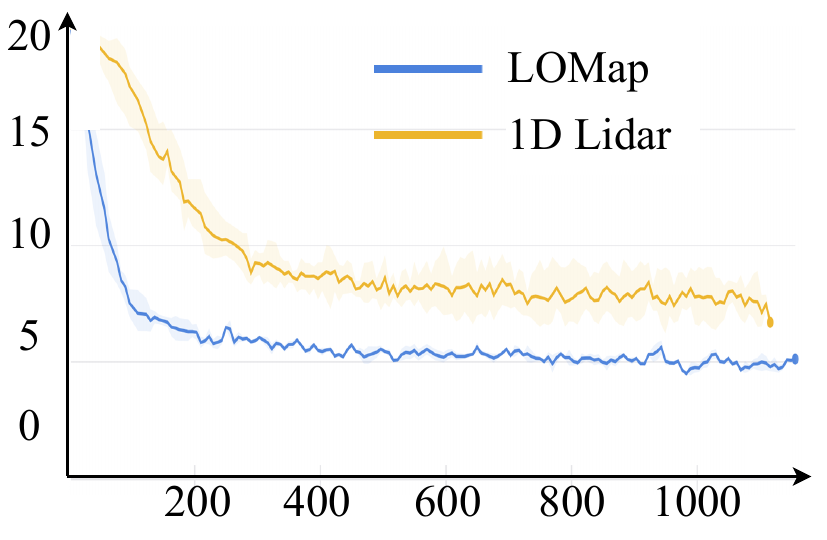}
	}
	\caption{Training curves of Lidar data and LOMap in high-level sub-goal generation.}
	\label{fig:image_LOMap_training_input}
	\vspace{-1em}
\end{figure}

To assess the enhancement in navigation performance afforded by the incorporation of LOMaps, we compared the use of raw lidar data with that of LOMaps as inputs for lidar information. Fig. \ref{fig:image_LOMap_training_input_return} and \ref{fig:image_LOMap_training_input_loss} demonstrate that the approach utilizing LOMaps as input achieves a higher cumulative reward more rapidly and reaches the goal with fewer steps. In conclusion, incorporating LOMaps as state inputs allows the robot to develop a more effective policy, facilitating long-term planning and making it particularly advantageous for sub-goal generation.

\subsubsection{Reward Function for Sub-Goal Generation}



\begin{figure}[ht]
	\centering
	\includegraphics[width=0.9\linewidth]{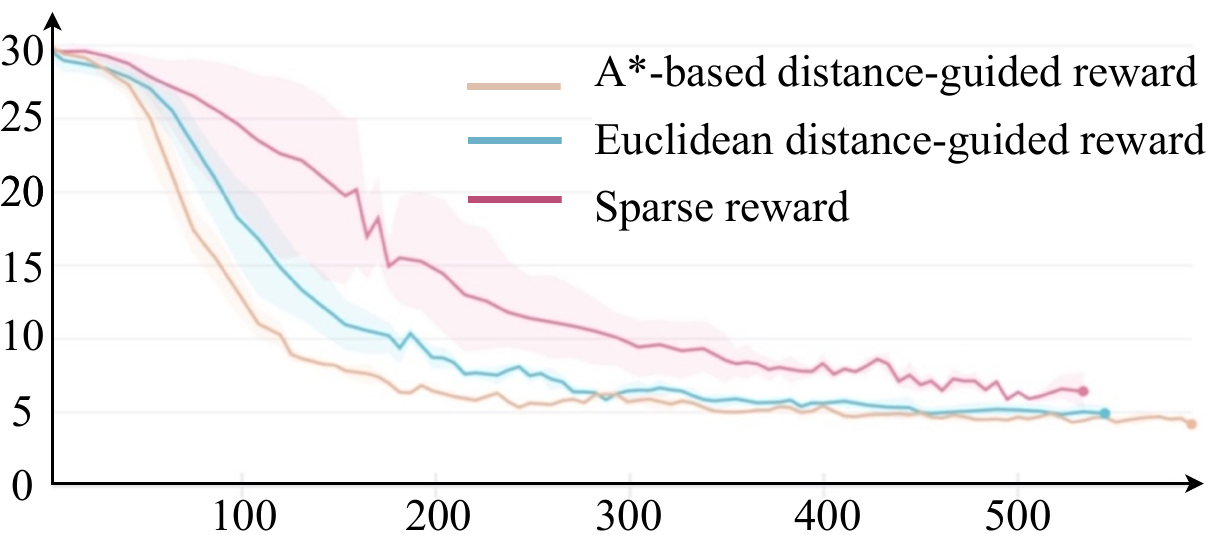}
	\caption{The average number of steps reached under different guidance reward.}
	\label{fig:image_high-level_reward_guidance}
\end{figure}

For the reward \( r_{\text{dist}} \) defined in (\ref{e:high-level_reward_dist}), we employ distance guidance based on the A* algorithm. This approach is contrasted with guidance based on Euclidean distance and a sparse reward scheme, where the reward is set to \( r = 0 \). As depicted in Fig. \ref{fig:image_high-level_reward_guidance}, utilizing A*-based distance guidance facilitates the robot's acquisition of skills like navigating through doorways and around obstacles, prevents entrapment in local minima area, and allows for goal attainment in a more efficient number of steps.

\subsubsection{Scenarios for for Sub-Goal Generation}

We assess the high-level sub-goal generation policy based on navigation distance, the presence of local minimum areas, and other conditions. (See the supplementary materials for details)

\subsubsection{Dynamic Obstacle Density for Low-level Policy}

We assess the performance of the low-level motion planning policy by varying the density of dynamic obstacles. As evidenced in Table \ref{tab:ablation_exp_Number-Pedestrians}, the policy, which is trained by SRL, facilitates the robot's navigation through dynamic environments with a sparse population. However, it encounters challenges in scenarios characterized by dense pedestrian traffic, occasionally resulting in collisions. This is primarily attributed to the robot's tendency to attempt navigating through the midst of groups of people moving in concert. Consequently, to mitigate the risk of unnecessary collisions, we incorporate post-processing techniques during the physical deployment of the navigation policy. Specifically, we implement a mechanism to preemptively suppress speed commands that lead to collisions with pedestrians.

\begin{table}[ht]
	\centering
	\setlength\tabcolsep{8pt}
	\begin{threeparttable}  
		\caption {Results of low-level policy in dynamic environments.}
		\label{tab:ablation_exp_Number-Pedestrians}
		\begin{tabular}{cccccc}
			\toprule
			Number of Pedestrians & SR  & SRN & CT & SPL   & SNT   \\
			\midrule
			0                     & 100 & 100 & 0  & 93.62 & 90.1  \\
			5                     & 100 & 100 & 0  & 91.05 & 90.17 \\
			10                    & 100 & 98  & 2  & 85.78 & 77.73 \\
			15                    & 100 & 96  & 7  & 83.25 & 72.21\\
			\bottomrule
		\end{tabular}
		\begin{tablenotes}
			\footnotesize
			\scriptsize
			\item[$^*$] We utilize a 30 $\times$ 30 meter test site for our experiments. The test scenarios are designed across four difficulty levels, corresponding to 0, 5, 10, and 15 pedestrians, respectively. The start and goal for the robot are randomly determined, with the Euclidean distance between them consistently exceeding 10 meters. We let the robot's navigation path pass through or accompany the motion trajectory of dynamic pedestrians.
		\end{tablenotes}
	\end{threeparttable}
\end{table}

\subsection{Comparison Experiments}

\subsubsection{Static Obstacle Scene}




We first simulate robot delivery or food delivery in static environments in office and home scenarios. The static scene navigation performance test is mainly to evaluate the mobile robot's ability to escape from local minima area and drive in a straight line. We use the low-level policy based on proximal policy optimization (PPO) \cite{schulman2017proximal} and CPO and the navigation policy based on HRL to test the navigation performance.

\begin{figure}[ht]
	\centering
	\subfigure[First office scene path set.]
	{
		\label{fig:image_simulation_path_results_static_office_1}
		\includegraphics[width=0.47\linewidth]{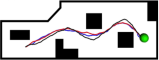}
	}
	\hspace{-2.5mm}
	\subfigure[Second office scene path set.]
	{
		\label{fig:image_simulation_path_results_static_office_2}
		\includegraphics[width=0.47\linewidth]{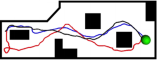}
	}
	\hspace{-2.5mm}
	\subfigure[First home scene path set.]
	{
		\label{fig:image_simulation_path_results_static_home_1}
		\includegraphics[width=0.47\linewidth]{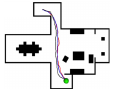}
	}
	\hspace{-2.5mm}
	\subfigure[Second home scene path set.]
	{
		\label{fig:image_simulation_path_results_static_home_2}
		\includegraphics[width=0.47\linewidth]{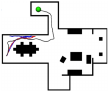}
	}
	\caption{Simulation results of paths in the static scenarios. The red, blue, and black curves represent the robot paths obtained from the low-level policies using the PPO and CPO algorithms, and the HRL-based navigation policy, respectively.}
	\label{fig:image_simulation_path_results_static_office}
	\vspace{-1em}
\end{figure}

\begin{table}[ht]
	\footnotesize
	\centering
	\setlength\tabcolsep{3.5pt}
	\begin{threeparttable}  
		\caption {Simulation path results in static scenes.}
		\label{tab:Simulation_path_results_in_static_scenes}
		\begin{tabular}{cccccccc}
			\toprule
			\multirow{2}{*}{Scenerio} & \multirow{2}{*}{Index} & \multirow{2}{*}{Policy} & \multicolumn{5}{c}{Metrics}                                                   \\
			\cmidrule(l){4-8}
			&                        &                         & SR           & SRN          & CT            & SPL            & SNT            \\
			\midrule
			\multirow{6}{*}{Office}   & \multirow{3}{*}{1st}   & Low-Level (PPO)         & 100          & 93           & 1.27          & 87.49          & 80.91          \\
			&                        & Low-Level (CPO)         & 100          & 100          & 0             & 92.26          & 85.12          \\
			&                        & HRL                     & \textbf{100} & \textbf{100} & \textbf{0}    & \textbf{92.56} & \textbf{87.66} \\
			\cmidrule(l){2-8}
			& \multirow{3}{*}{2nd}   & Low-Level (PPO)         & 95           & 0            & 39.30         & 73.52          & 59.95          \\
			&                        & Low-Level (CPO)         & 98           & 42           & 15.34         & 86.59          & 69.75          \\
			&                        & HRL                     & \textbf{100} & \textbf{83}  & \textbf{3.37} & \textbf{91.77} & \textbf{73.19} \\
			\midrule
			\multirow{6}{*}{Home}     & \multirow{3}{*}{1st}   & Low-Level (PPO)         & 100          & 85           & 0.32          & \textbf{92.33} & \textbf{90.57} \\
			&                        & Low-Level (CPO)         & 100          & 100          & 0             & 90.52          & 83.23          \\
			&                        & HRL                     & \textbf{100} & \textbf{100} & \textbf{0}    & 90.87          & 84. 97         \\
			\cmidrule(l){2-8}
			& \multirow{3}{*}{2nd}   & Low-Level (PPO)         & 0            & 0            & 63.15         & 0              & 0              \\
			&                        & Low-Level (CPO)         & 0            & 0            & 23.72         & 0              & 0              \\
			&                        & HRL                     & \textbf{97}  & \textbf{92}  & \textbf{0.43} & \textbf{91.16} & \textbf{84.95}\\
			\bottomrule
		\end{tabular}
	\end{threeparttable}
\end{table}

As shown in Table \ref{tab:Simulation_path_results_in_static_scenes}, in simple office and home scenarios like those in Fig. \ref{fig:image_simulation_path_results_static_office_1} and Fig. \ref{fig:image_simulation_path_results_static_home_1}, the three navigation policies perform similarly, except that the PPO-based low-level policy has a higher collision probability when navigating corners. In scenarios with narrow channels, such as those in Fig. \ref{fig:image_simulation_path_results_static_office_2} and Fig. \ref{fig:image_simulation_path_results_static_home_2}, the performance of all three policies is significantly reduced, resulting in tortuous paths and numerous collisions. The long path lengths and complex obstacles render navigation by the PPO-based and CPO-based low-level policies to the goal impossible. However, the HRL-based navigation policy excels over the PPO-based and CPO-based policies, particularly in the local minimum area depicted in Fig. \ref{fig:image_simulation_path_results_static_home_2}. It can successfully escape and guide the robot to the goal. These results demonstrate that the HRL-based navigation policy possesses strong generalization capabilities, enabling efficient navigation in indoor scenes with local minimum areas that have not been seen during policy training.

\subsubsection{Dynamic Obstacle Scene}

The dynamic navigation experiment modeles the conditions of a robot in crowded environments like restaurants, primarily assessing the robot's obstacle avoidance and rapid response abilities within dense crowds. Concurrently, to evaluate the generalization performance of the HRL-based navigation policy, we employ the policy model trained in the static experiment without additional fine-tuning.

\begin{figure}[ht]
	\centering
	\subfigure[]
	{
		\label{fig:dynamic_path-1}
		\includegraphics[width=0.4\linewidth]{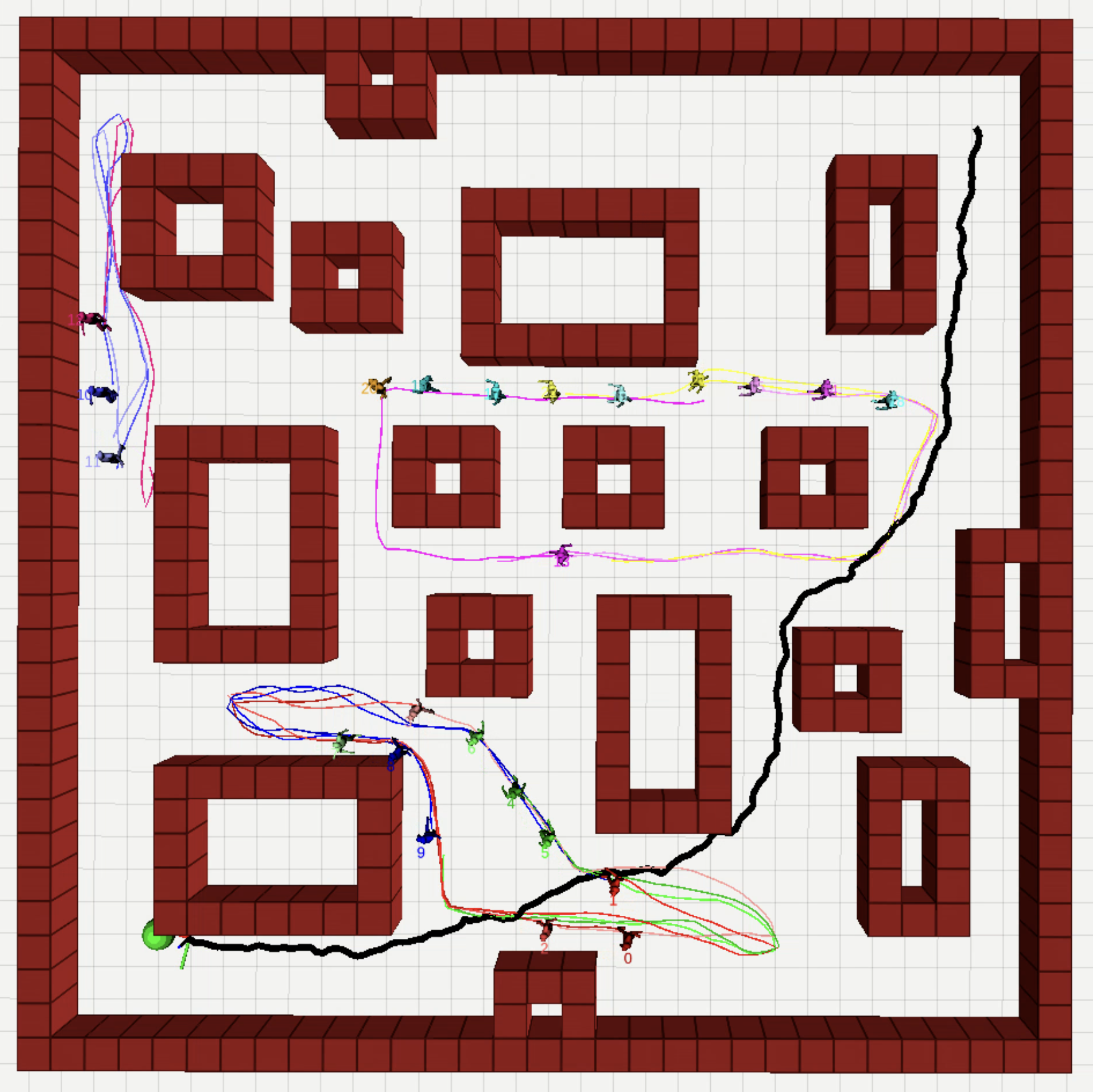}
	}
	\hspace{-2.5mm}
	\subfigure[]
	{
		\label{fig:dynamic_path-2}
		\includegraphics[width=0.4\linewidth]{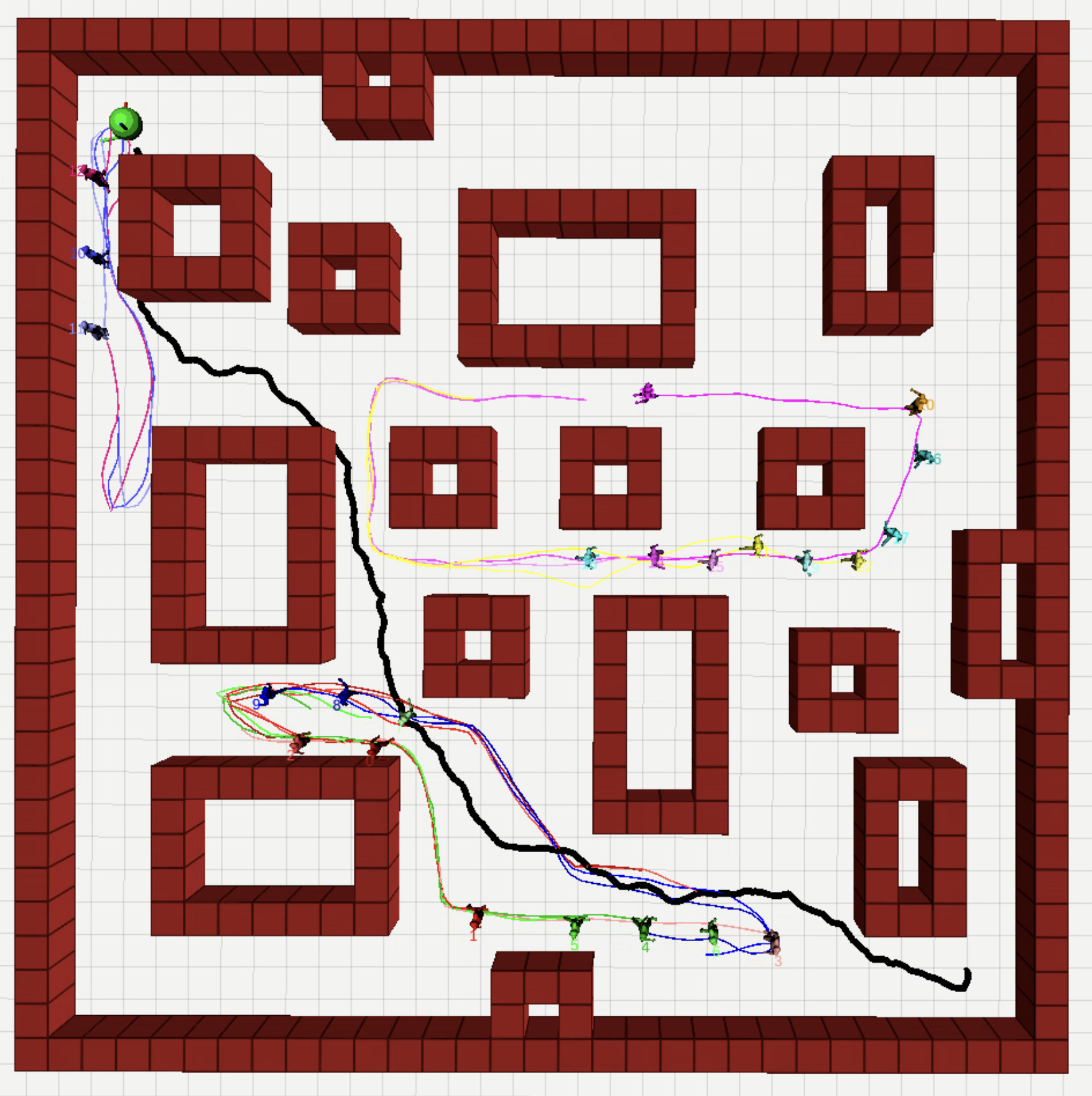}
	}
	\caption{Robot path in the dynamic experiment (black). The green dot is the robot's goal, and the other colored dots represent pedestrians and their trajectories. The simulation environment models the layout of a restaurant, with a total of 23 dynamically simulated pedestrians.}
	\label{fig:image_simulation_path_results_dynamic}
	\vspace{-1em}
\end{figure}


\begin{table}[ht]
	\footnotesize
	\centering
	\setlength\tabcolsep{10pt}
	\begin{threeparttable}  
		\caption {Simulation results in dynamic scenarios.}
		\label{tab:map_test_agent_density}
		\begin{tabular}{ccccc}
			\toprule
			No.&Velocity of Pedestrians & SR  & SRN & CT   \\
			\midrule
			1&0.5 m/s      & 100 & 100 & 0    \\
			2&0.8 m/s      & 97  & 90  & 0.15 \\
			3&1.2 m/s      & 95  & 78  & 0.75 \\
			\bottomrule
		\end{tabular}
				\begin{tablenotes}
					\footnotesize
					\scriptsize
					\item[$^*$] The number of pedestrians is 23.
				\end{tablenotes}
	\end{threeparttable}
\end{table}

Fig. \ref{fig:image_simulation_path_results_dynamic} illustrates that the robot follows a nearly optimal path and timely avoids crowds. 
Table \ref{tab:map_test_agent_density} shows that in dynamic scenes with slow pedestrians, the HRL-based navigation policy maintains a low collision rate for the robot without post-processing. However, as pedestrian speed increases, the robot's success rate in collision-free navigation gradually decreases, while its collision rate significantly rises.


\section{Validation Experiments}
\label{sec:Validation_Experiments}

\begin{figure}[ht]
	\centering
	\subfigure[TurtleBot3-Burger]
	{
		\label{fig:image_turtlebot}
		\includegraphics[height=1.3in]{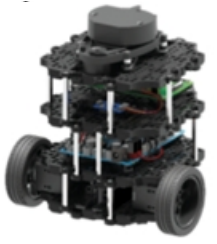}
	}
	\hspace{-2.5mm}
	\subfigure[Real-world navigation environment]
	{
		\label{fig:image_exp_env}
		\includegraphics[height=1.3in]{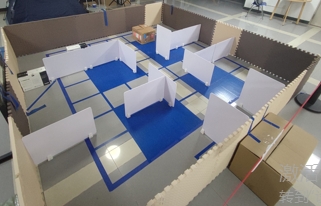}
	}
	\caption{Physical robot and environment.}
	\label{fig:image_exp_setting}
	\vspace{-1em}
\end{figure}

As shown in Fig. \ref{fig:image_exp_setting},
this paper conducts physical experiments using the TurtleBot3-Burger robot, equipped with a single-line lidar. Based on the actual hardware parameters of the TurtleBot3, we set the maximum linear velocity output of the neural network to 0.22 $m/s$. The experimental scenes, which include static and dynamic obstacles, are constructed using plastic plates.

\begin{figure}[ht]
	\centering
	\subfigure[]
	{
		\label{fig:image_validation_path_results_static-1}
		\includegraphics[width=0.305\linewidth]{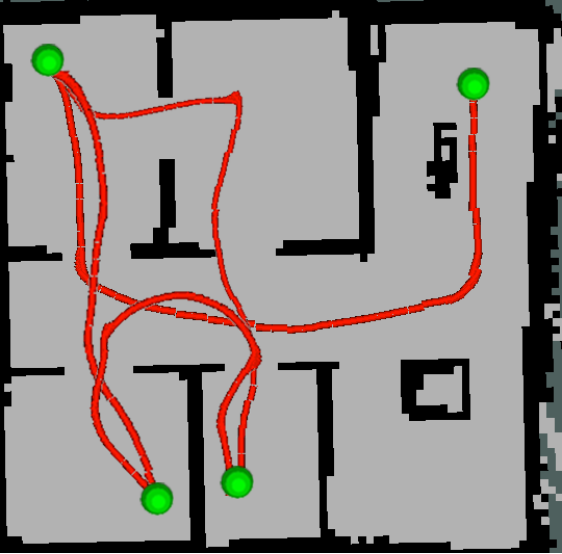}
	}
	\hspace{-2.5mm}
	\subfigure[]
	{
		\label{fig:image_validation_path_results_static-2}
		\includegraphics[width=0.305\linewidth]{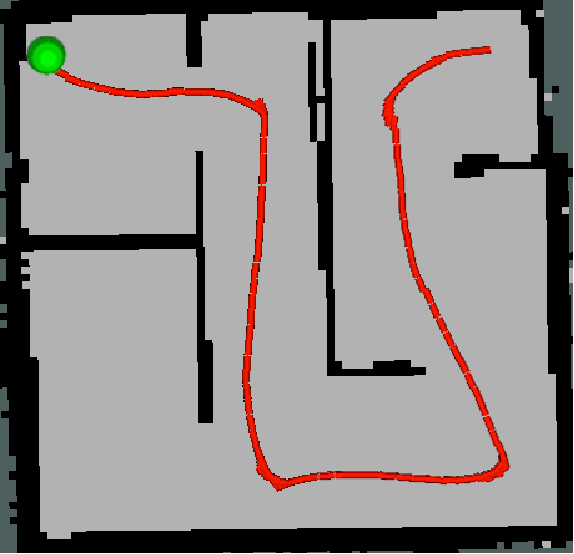}
	}
	\hspace{-2.5mm}
	\subfigure[]
	{
		\label{fig:image_validation_path_results_dynamic}
		\includegraphics[width=0.305\linewidth]{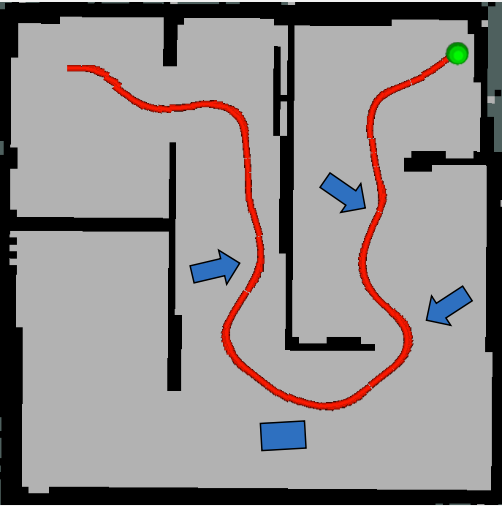}
	}
	\caption{TurtleBot3's navigation path in the physical experiment (red). The green dot indicates the robot's goal. (a) Static scene with four goals. (b) Static scene for long-distance navigation. (c) Dynamic Scene. Arrows indicate external dynamic interference.}
	\label{fig:image_validation_path_results_static}
	\vspace{-1em}
\end{figure}

We initially conduct 30 navigation tests in the scenario depicted in Fig. \ref{fig:image_validation_path_results_static-1} with four goals and 10 tests in the long-distance navigation scenario shown in Fig. \ref{fig:image_validation_path_results_static-2}. Despite hardware parameter variations, the policy model consistently guides the robot to successfully reach goals in both local minimum areas and long-distance navigation scenarios without aimless movement. Subsequently, we introduce dynamic obstacles into the scenario from Fig. \ref{fig:image_validation_path_results_static-2} to assess dynamic navigation performance, with results presented in Fig. \ref{fig:image_validation_path_results_dynamic}. The results indicate that TurtleBot3 can timely adjust its direction to avoid interfering obstacles, resulting in no collisions. These experiments demonstrate that the HRL-based navigation policy performs effectively on physical robots and exhibits strong generalization capabilities.

%
%
%
%
%


\section{Conclusion}
\label{sec:conclusions}

This paper presents a robust mapless navigation framework leveraging HRL to effectively 
navigate local minima area in indoor environments. The high-level policy's dynamic sub-goal updates, coupled with an enhanced obstacle encoding method at the low level, ensure real-time adaptability and improved environmental perception. 
Through extensive simulations across diverse settings, including offices, homes, and restaurants, our HRL framework has consistently demonstrated superior performance in navigating static and dynamic obstacles alike. The successful physical deployment on a TurtleBot3 robot further underscores the practical applicability and reliability of our navigation policy.
While the HRL-based navigation policy performs well in various settings, it encounters occasional collisions in scenes with dense obstacles. Future enhancements to our algorithm, potentially aided by large language models, are planned to address these challenges.

\ifCLASSOPTIONcaptionsoff
\newpage
\fi

\end{document}